\documentclass[letterpaper]{article}
\usepackage[preprint]{aaai2027}
\usepackage[hyphens]{url}
\usepackage{graphicx}
\usepackage{natbib}
\usepackage{caption}
\usepackage{listings}
\usepackage{placeins}
\usepackage{algorithm}
\usepackage{booktabs}
\usepackage{tabularx}
\title{\parbox{\textwidth}{\centering HookEvo: A Framework for Coordinated Plugin Evolution\\on Hook-Enabled Agent Runtimes}}
\author{\parbox{\textwidth}{\raggedright\large\bfseries \mbox{Shengze Huang\textsuperscript{1,7}}, \mbox{Yanyu Mao\textsuperscript{2}}, \mbox{Xiaofei Yu\textsuperscript{3}}, \mbox{Baoshen Guo\textsuperscript{4,8}}, \mbox{Xiaorui Wang\textsuperscript{5}}, \mbox{Zhiwen Wu\textsuperscript{5}}, \mbox{Sheng Wu\textsuperscript{5}}, \mbox{Fukang Wen\textsuperscript{1}}, \mbox{Chengxu Wu\textsuperscript{5}}, \mbox{Linhao Li\textsuperscript{6}}, \mbox{Kailai Sun\textsuperscript{4,8}}, \mbox{Baotong Mi\textsuperscript{7}}}}
\affiliations{\parbox{\textwidth}{\raggedright\normalsize \mbox{\textsuperscript{1}\,Tsinghua University}\hspace{1em plus .15em minus .1em} \mbox{\textsuperscript{2}\,Nanyang Technological University}\hspace{1em plus .15em minus .1em} \mbox{\textsuperscript{3}\,Renmin University of China}\hspace{1em plus .15em minus .1em} \mbox{\textsuperscript{4}\,Massachusetts Institute of Technology}\hspace{1em plus .15em minus .1em} \mbox{\textsuperscript{5}\,Peking University}\hspace{1em plus .15em minus .1em} \mbox{\textsuperscript{6}\,Southeast University}\hspace{1em plus .15em minus .1em} \mbox{\textsuperscript{7}\,Renlab AI}\hspace{1em plus .15em minus .1em} \mbox{\textsuperscript{8}\,Singapore-MIT Alliance for Research and Technology}}}
\usepackage{amsmath}
\usepackage{amssymb}
\usepackage{array}
\usepackage{colortbl}
\usepackage{xcolor}
\definecolor{ourrow}{HTML}{EAF3FA}
\definecolor{grouphead}{HTML}{E9EFF4}
\begin{document}
\maketitle

\begin{abstract}
Repairing failures in tool-using agents can require coordinated changes to how the harness interprets tool receipts, assembles context, and checks actions. We introduce HookEvo, a framework that uses an existing agent runtime's lifecycle hooks to evolve task-specific harnesses. We instantiate HookEvo on DeepSeek Harness. In each round, a separate improver diagnoses failures from execution traces and proposes a candidate. HookEvo evaluates the candidate against the current harness on matched tasks and carries the retained version into the next round. A single candidate can jointly edit executable plugin handlers and the state they share within a task, allowing a tool receipt to inform context and control a later action. Using a frozen Qwen3-4B executor, the selected harnesses lead the reported Qwen3-4B comparisons, achieving 95.41\% on ALFWorld, 70.56\% on DBBench, 48.17\% on WebShop, and 41.67\% on OSBench. Across three improvers and three edit scopes, from skill text to single-component and coordinated executable edits, GPT-6.1 Sol shows the largest average gain from the broadest scope.
\par\noindent Our code is available at \pdfstartlink attr {/Border [0 0 0]} user {/Subtype /Link /A << /S /URI /URI (https://huangshengzebluesky.github.io/HookEvo/) >>}{\color{blue}\url{https://huangshengzebluesky.github.io/HookEvo/}}\pdfendlink.
\end{abstract}

\section*{1 Introduction}
\begin{figure*}[!t]
\centering
\includegraphics[width=\linewidth]{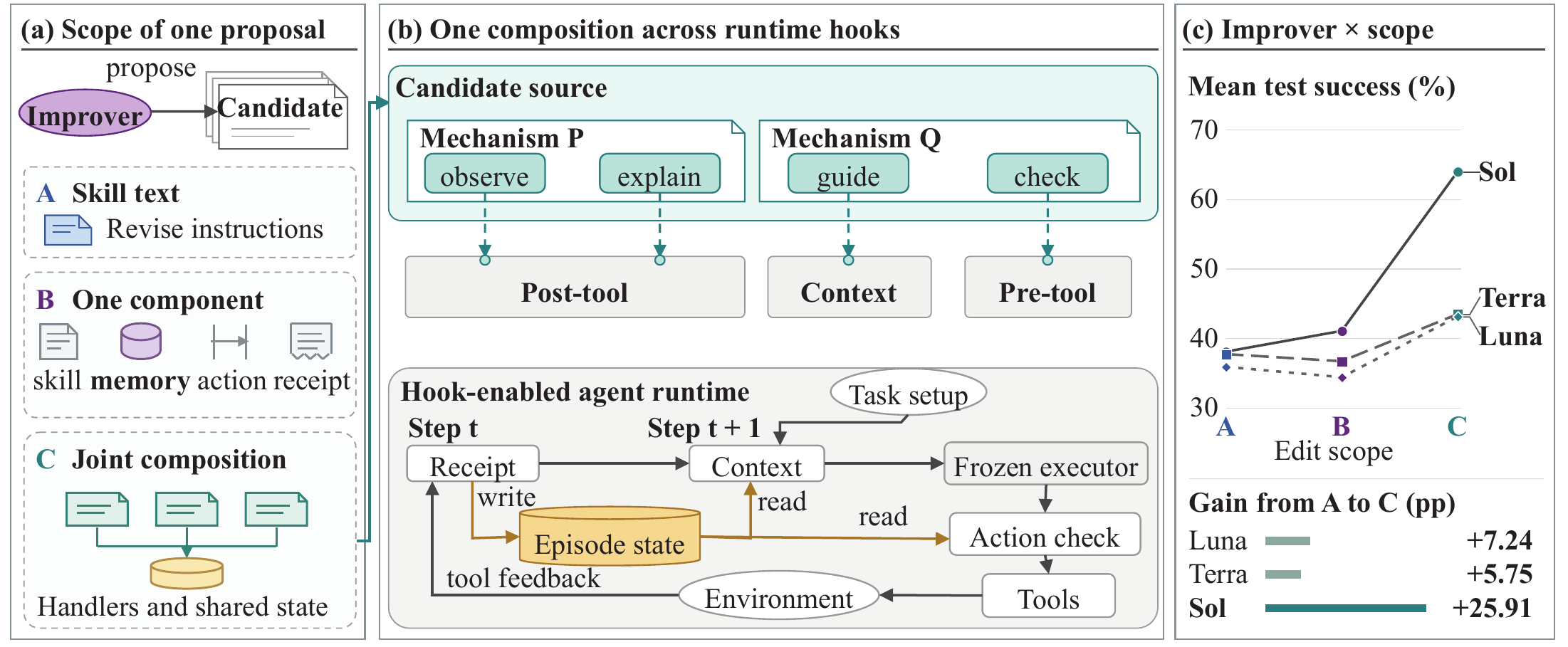}
\caption{Edit scope and improver performance. (a) One proposal can revise skill text (A), one independent component (B), or a composition of mechanisms that share episode state (C). (b) In the illustrated C candidate, one mechanism processes a tool receipt and another uses the resulting state to guide context assembly and check the next action. (c) Mean test success across four benchmarks and the observed gain from A to C in percentage points (Table~\ref{tab:scope-results}).}
\label{fig:space}
\end{figure*}

A tool-using agent relies on a harness to assemble the model's context and mediate tool calls~\citep{react,sweagent,codeact}. After a tool returns a result, the harness determines what information reaches the model at the next step and whether a subsequent tool call can proceed. Changing how the harness uses this feedback can therefore change the agent's behavior even when the model weights are fixed.

Prior work uses execution feedback to optimize different parts of an agent harness. SkillOpt proposes edits to a skill document and retains revisions that improve performance on a validation set~\citep{skillopt}. AFlow searches over workflows expressed in code, using execution results to guide changes to their operators and control flow~\citep{aflow}. PluginRSI improves individual plugins, adds the resulting variants to a library, and recombines them into candidate harnesses~\citep{pluginrsi}. These choices determine which changes an improver can propose together and evaluate as a single candidate.

The scope of a candidate matters when a repair requires changes that only work together. Recording a fact from a tool result may leave the task score unchanged if no later action check reads it. Revising the check alone may be equally ineffective if the harness never records the fact it needs. Evaluated separately, either edit may be discarded before the other is proposed. Evaluating them together lets their combined effect count toward retention, but the improver must first recognize the dependency in execution traces and revise both parts consistently.

To address this challenge, we introduce HookEvo, a framework that lets an improver revise interdependent parts of an agent harness in a single candidate. We implement HookEvo on DeepSeek Harness (DSH)~\citep{deepseek-harness}, whose lifecycle hooks expose tool receipts, context assembly, and action checks. Given execution traces and the current harness source, the improver can jointly change what one handler records in episode state and how another uses that state at a later step. HookEvo evaluates the resulting candidate against its parent on matched tasks, so the retention decision reflects the combined effect of the edits. The retained harness becomes the parent for the next round, and the measured outcome informs the next proposal.

We evaluate HookEvo through three research questions.

\noindent\textbf{RQ1.} \emph{Across task scenarios, how do the harnesses evolved by HookEvo perform relative to prior methods using the same executor?}\par
\noindent\textbf{RQ2.} \emph{How does the benefit of allowing coordinated edits in a single candidate vary across improvers?}\par
\noindent\textbf{RQ3.} \emph{How do the improver and the edit scope shape the mechanisms in the evolved harnesses?}

With a frozen Qwen3-4B executor, we compare the harnesses evolved by Sol under the coordinated edit scope (C) with a same-executor Life-Harness rerun and published Qwen3-4B results across four benchmarks. We then compare test success for Luna, Terra, and Sol under three edit scopes: skill text (A), one independent component (B), and coordinated edits to handlers and shared state (C), as illustrated in Figure~\ref{fig:space}. To examine the mechanisms behind these results, we compare the selected harnesses across improvers under C and across edit scopes for Sol.

Our contributions are threefold.
\begin{itemize}
\item \textbf{A unified representation and evolution method.} HookEvo represents harness edits as changes to plugin content, code, and composition on an existing hook-enabled runtime. The improver can revise both the handlers that write shared state and those that read it. The controller evaluates their combined effect on matched tasks.
\item \textbf{A study of edit scope and improver capability.} We cross three edit scopes with three improvers at high reasoning effort. Moving from A to C increases mean test success across four benchmarks by 25.91 percentage points for Sol, 7.24 for Luna, and 5.75 for Terra.
\item \textbf{Benchmark results and reproducible artifacts.} With a frozen Qwen3-4B executor, the selected Sol-C harnesses lead the reported Qwen3-4B comparisons on ALFWorld (95.41\%), DBBench (70.56\%), WebShop (48.17\%), and OSBench (41.67\%). We provide the harnesses, checkpoint results, evolution histories, and candidate-generation interfaces.
\end{itemize}

\section*{2 Related Work}
\begin{figure*}[!t]
\centering
\includegraphics[width=\linewidth]{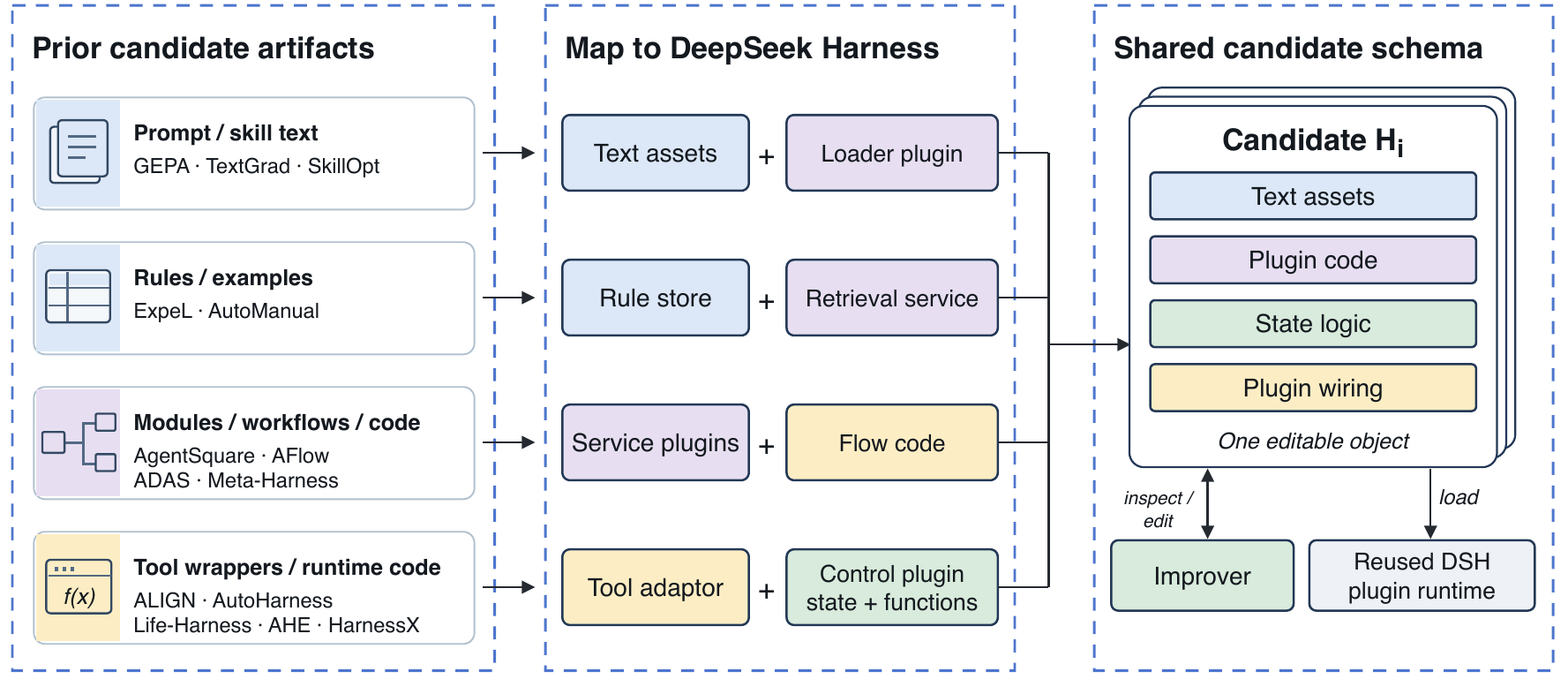}
\caption{Mapping prior candidate artifacts to a shared HookEvo representation. The left column groups the artifacts optimized by prior methods. The middle column gives possible implementations using DeepSeek Harness (DSH) plugins and services. The right column shows a candidate whose text, plugin code, state logic, and wiring can be inspected and edited together by the improver and loaded into the existing DSH runtime.}
\label{fig:related-candidates}
\end{figure*}

Figure~\ref{fig:related-candidates} maps prior candidate artifacts to possible implementations on DSH. For example, an improver can edit skill text loaded into context or an action filter that runs before a tool call. The host installs plugins through \texttt{ctx.plugin(module, config)}, and the installed wrapper connects candidate handlers to DSH lifecycle events.

\paragraph{Prompt and program optimization.}
OPRO searches instructions using the scores of earlier candidates, while ProTeGi converts minibatch errors into textual update directions~\citep{opro,protegi}. DSPy expresses language-model pipelines as optimizable programs; MIPRO jointly searches their instructions and demonstrations~\citep{dspy,mipro}. TextGrad propagates language feedback through computation graphs~\citep{textgrad}, and GEPA combines trajectory reflection with Pareto-based candidate search~\citep{gepa}. SkillOpt optimizes a reusable skill through bounded edits, validation, and rejected-edit feedback~\citep{skillopt}. These methods establish feedback-guided optimization over language artifacts and modular programs. Our study examines how an improver uses broader permissions to construct executable harness mechanisms.

\paragraph{Feedback, experience, and reusable skills.}
Self-Refine improves an output through iterative feedback, and Reflexion carries verbal feedback into subsequent trials~\citep{selfrefine,reflexion}. ExpeL extracts reusable insights from task experience; AutoGuide selects context-aware guidelines; AutoManual builds environmental rules and instruction manuals~\citep{expel,autoguide,automanual}. Voyager accumulates executable skills in a persistent library~\citep{voyager}. HookEvo uses episode state for a different purpose: sharing facts from the current task across handlers.

\paragraph{Agent architecture and workflow search.}
ADAS uses a meta agent to write agent programs from an archive of previous designs~\citep{adas}. AgentSquare evolves and recombines planning, reasoning, tool-use, and memory modules~\citep{agentsquare}. AFlow searches code-represented workflows by revising operators and their composition~\citep{aflow}.

\paragraph{Runtime interfaces and harness evolution.}
SWE-agent demonstrates the importance of agent-computer interfaces, and CodeAct uses executable code as a compositional action space~\citep{sweagent,codeact}. ALIGN generates environment interfaces; AutoHarness synthesizes code harnesses from environment feedback~\citep{align,autoharness}. Life-Harness organizes adaptation across runtime lifecycle layers~\citep{life}. Meta-Harness gives its proposer access to prior source, scores, and traces~\citep{meta}; AHE connects component, trajectory, and decision observability~\citep{ahe}; HarnessX evolves typed, composable harness primitives~\citep{harnessx}. These systems establish executable runtime behavior as an optimization target.

The right column of Figure~\ref{fig:related-candidates} shows how HookEvo combines editable text, plugin code, state logic, and plugin wiring into one candidate. A candidate can revise instructions, replace one component, or jointly revise handlers and the episode state they share.

\section*{3 Method}
\label{sec:method}
\begin{figure*}[!t]
\centering
\includegraphics[width=\textwidth]{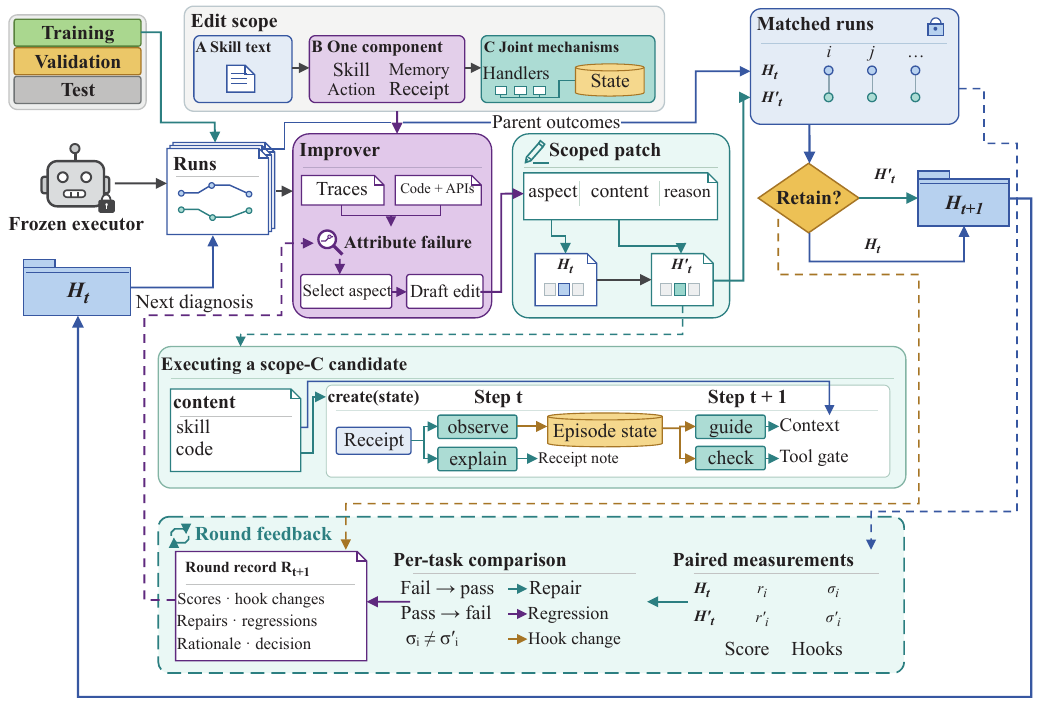}
\caption{One round of HookEvo. Parent runs supply traces for the improver and baseline outcomes for paired evaluation. A scoped patch yields one candidate. Under scope C, its handlers can share state across execution steps within a task. The controller compares the two harnesses on matched runs, carries the retained source into the next round, and returns measured changes and its decision to the improver.}
\label{fig:method-overview}
\end{figure*}

HookEvo optimizes a harness $H=(\lambda,G)$ around a frozen executor. Here $\lambda$ contains skill instructions and configuration, and $G$ defines the executable handlers, their state, and their dependencies. For a task distribution $D$ and normalized task score $r_H$, the objective is to minimize $\mathcal L_D(H)=1-\mathbb E_{x\sim D,\omega}[r_H(x;\omega)]$, where $\omega$ denotes execution randomness. Figure~\ref{fig:method-overview} follows one update from the current harness $H_t$ to the version retained for the next round.

\subsection*{3.1 Composable Candidate Space}
Let $\mathcal E_k(H)$ denote the edits that a single proposal may make to $H$ under scope $k$. Scope A exposes the skill text. Scope B exposes four independent components (skill, memory, action check, and receipt interpretation) and permits a proposal to replace one. Scope C permits a skill edit or a joint replacement of skill text and executable plugin code. The code defines an ordered collection of mechanisms whose handlers can share episode state. A single proposal can therefore change what information is recorded and how later handlers use it.

\subsection*{3.2 Observation, Attribution, and Candidate Construction}
At round $t$, the controller runs $H_t$ on training batch $B_t$. The execution traces provide evidence $\mathcal T_t$ for the improver, while the controller keeps the task scores and emitted hook outputs as the baseline for this round. The improver $M$ examines those traces alongside the parent source, the available runtime interfaces $A_t$, and feedback $\mathcal R_t$ from the previous round. It relates a failure in the traces to behavior that the current harness can change:
\begin{equation}
g_t=\operatorname{Diagnose}_M(H_t,A_t,\mathcal T_t,\mathcal R_t).
\label{eq:direction}
\end{equation}
The diagnosis $g_t$ serves as a repair direction in discrete program space~\citep{protegi,textgrad}. The edit scope determines which replacements can implement that repair:
\begin{align}
e_t &\sim q_M(\cdot\mid H_t,g_t,\mathcal R_t,\mathcal E_k(H_t)),\nonumber\\
H'_t&=\operatorname{Apply}(H_t,e_t).
\label{eq:proposal}
\end{align}
Attribution and editing take place in one improver call. The response names the target (\texttt{aspect}), supplies its replacement (\texttt{content}), and gives the rationale (\texttt{reason}). The controller copies $H_t$ and installs the replacement to form $H'_t$; the candidate inherits all other parts of the parent. If assembly fails, the controller retains $H_t$ and returns the error as feedback.

For a scope-C composition, DeepSeek Harness (DSH)~\citep{deepseek-harness} passes fresh episode state to \texttt{create(state)} at task setup. On a tool receipt, \texttt{observe} can update the state and \texttt{explain} can annotate the receipt. At a later step, \texttt{guide} can read the state to augment the context alongside the skill text, and \texttt{check} can read it before a proposed tool call proceeds. These handlers are optional. The controller evaluates the installed composition as a unit. Retaining the code preserves these bindings across rounds, while each new task starts with fresh state.

\subsection*{3.3 Paired Evaluation, Retention, and Feedback}
The controller runs $H'_t$ with the same frozen executor and pairs each outcome with the parent outcome for the same task, trial, and seed. For matched runs $i\in\mathcal I_t$, the mean paired score change is
\begin{equation}
\widehat\Delta_t=\frac{1}{|\mathcal I_t|}\sum_{i\in\mathcal I_t}
\big[r_i(H'_t)-r_i(H_t)\big].
\label{eq:paired}
\end{equation}
For each pair, it also compares the emitted hook outputs, denoted by $\sigma_i$ for the parent and $\sigma'_i$ for the candidate in Figure~\ref{fig:method-overview}. A transition from failure to success counts as a repair, and the reverse counts as a regression.

The candidate becomes $H_{t+1}$ if it improves the benchmark's training objective, changes the hook outputs on at least one matched run, and, when a validation split is available, maintains or improves the validation objective. Otherwise $H_{t+1}=H_t$. In either case, the next improver call receives $\mathcal R_{t+1}$ with the edit rationale, before-and-after scores, observed hook changes, counts of repairs and regressions, and the retention decision. Held-out test evaluation follows the completed search. Appendix~A gives the algorithm and runtime interfaces, and Appendix~B describes the A/B/C experimental setup.

\section*{4 Experiments}
\begin{figure*}[!t]
\centering
\includegraphics[width=\textwidth]{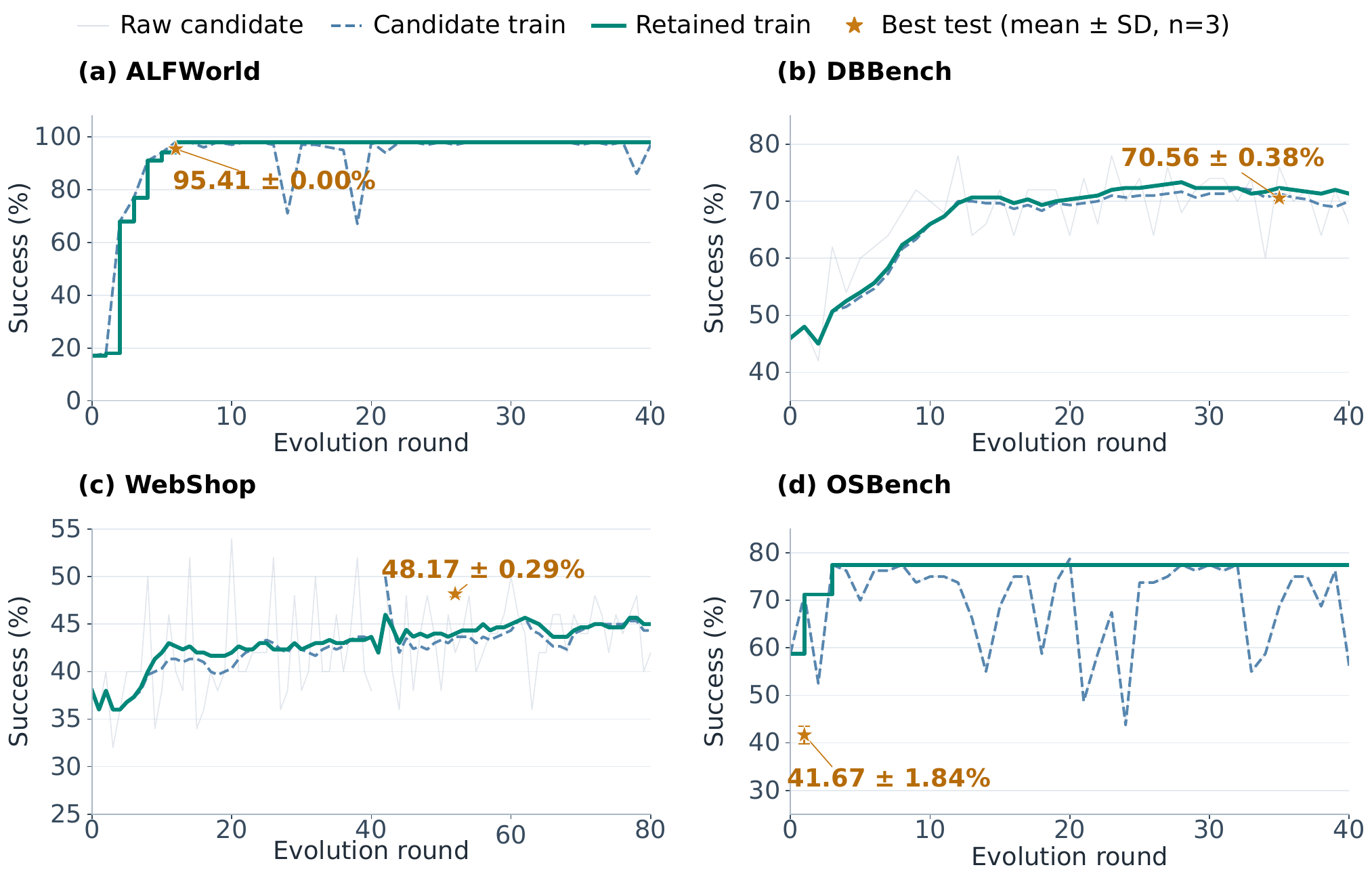}
\caption{Sol-C search histories (Table~\ref{tab:main-results}): 40 rounds for ALFWorld, DBBench, and OSBench; 80 across two WebShop stages. Gray shows candidate scores, dashed blue their trend, and green retained scores (six-round means for DBBench and WebShop). Stars mark peak test means with sample SD over three executions.}
\label{fig:evolution}
\end{figure*}
\subsection*{Experimental setup}

\noindent\textbf{Tasks and metric.} The same frozen Qwen3-4B-Instruct-2507 executor~\citep{qwen3} acts in four environments: ALFWorld household tasks~\citep{alfworld} (109 test tasks), DBBench database tasks (300), WebShop shopping tasks~\citep{webshop} (200), and OSBench shell tasks (144). DBBench and OSBench come from AgentBench~\citep{agentbench}. We report the percentage of held-out tasks completed under each benchmark's native success criterion. On WebShop, only a fully successful purchase counts. Table~\ref{tab:reference-splits} lists the reference task splits.

\noindent\begin{minipage}{\columnwidth}
\centering
{\small
\begin{tabular}{@{}lrrr@{}}
\toprule
Dataset & Train & Validation & Test \\
\midrule
ALFWorld & 300 & 0 & 109 \\
DBBench & 300 & 0 & 300 \\
WebShop & 100 & 0 & 200 \\
OSBench & 80 & 20 & 144 \\
\bottomrule
\end{tabular}\par}
\captionof{table}{Reference splits (distinct tasks).}
\label{tab:reference-splits}
\end{minipage}

\begin{table*}[!t]
\centering
\begingroup
\setlength{\tabcolsep}{1.4pt}
\renewcommand{\arraystretch}{1.08}
\small
\begin{tabularx}{\textwidth}{@{}p{0.32\textwidth}*{4}{>{\centering\arraybackslash}X}@{}}
\toprule
Method & ALFWorld & DBBench & WebShop & OSBench \\
\midrule
\rowcolor{grouphead}\multicolumn{5}{l}{\textbf{Results reported in prior work}} \\
Life-Harness~\citep{life} & 88.07 & \underline{65.00} & \underline{45.50} & \underline{39.60} \\
Bare~\citep{life} & 16.51 & 44.00 & 29.50 & 21.50 \\
ALIGN~\citep{align} & 28.36 & — & 18.00 & — \\
AgentSquare~\citep{agentsquare} & 29.36 & — & 18.50 & — \\
Meta-Harness (Life rerun)~\citep{meta} & 35.80 & 50.30 & 42.50 & 25.70 \\
AHE (Life rerun)~\citep{ahe} & \underline{89.00} & 57.30 & 39.00 & 37.50 \\
Harness-R1 (paper)~\citep{harness-r1} & 29.10 & 55.70 & 38.60 & — \\
\addlinespace[3pt]
\rowcolor{grouphead}\multicolumn{5}{l}{\textbf{Local evaluations}} \\
Life-Harness (local)~\citep{life} & 87.77$\pm$1.91 & 64.11$\pm$0.38 & 41.33$\pm$0.29 & 36.81$\pm$0.00 \\
\rowcolor{ourrow}\textbf{HookEvo (Sol, scope C)} & \textbf{95.41$\pm$0.00} & \textbf{70.56$\pm$0.38} & \textbf{48.17$\pm$0.29} & \textbf{41.67$\pm$1.84} \\
\bottomrule
\end{tabularx}
\endgroup
\smallskip
\begin{minipage}{\textwidth}\small
Test success (\%); local results show sample SD. Harness-R1 specifies Qwen3-4B only and tests 500 WebShop tasks (ours: 200). Life-Harness reports ALIGN and AgentSquare (Table 11) and reruns Meta-Harness and AHE (Table 1)~\citep{life}. Bold: best; underline: runner-up.
\end{minipage}
\par\normalsize
\caption{Reported and local benchmark results for Qwen3-4B-family executors.}
\label{tab:main-results}
\end{table*}

\noindent\textbf{Search and comparisons.} An external improver (Luna, Terra, or Sol, each at high reasoning effort) proposes harness edits from execution traces. A proposal may revise skill text (A), replace one independent component (B), or jointly revise handlers and shared state (C). Rotating pools use six 50-task batches, with one proposal per batch. We compare the resulting harnesses with published Qwen3-4B results and a local Life-Harness rerun using the same executor. Means weight benchmarks equally. Appendix~B gives the search budgets and evaluation protocols; Appendix~D defines the metrics.

\subsection*{4.1 RQ1: Benchmark Performance and Applicability}
The same search loop produces a separate harness for each environment using its native tools and scorer. Sol-C ranks first among the displayed Qwen3-4B results (Table~\ref{tab:main-results}). Its highest measured test means exceed the local Life-Harness rerun by 7.64, 6.45, 6.84, and 4.86 points on ALFWorld, DBBench, WebShop, and OSBench, respectively. Figure~\ref{fig:evolution} locates these checkpoints along each search.

\begin{table*}[!t]
\centering
\begingroup
\small
\renewcommand{\arraystretch}{1.10}
\begin{tabularx}{\textwidth}{@{}ll*{4}{>{\centering\arraybackslash}X}c@{}}
\toprule
Improver & Scope & ALFWorld & DBBench & WebShop & OSBench & Mean \\
\midrule
GPT-6 Luna & A & 23.85 & 53.00 & 32.33 & 34.26 & 35.86 \\
 & B & 14.07 & 54.56 & 31.33 & 37.50 & 34.36 \\
 & C & 48.62 & 57.00 & 30.67 & 36.11 & 43.10 \\
\addlinespace[3pt]
GPT-6.1 Terra & A & 26.91 & 53.78 & 30.50 & 39.58 & 37.69 \\
 & B & 29.97 & 53.78 & 29.33 & 33.56 & 36.66 \\
 & C & 52.91 & 54.00 & 32.83 & 34.03 & 43.44 \\
\addlinespace[3pt]
GPT-6.1 Sol & A & 27.22 & 52.89 & 34.33 & 37.73 & 38.04 \\
 & B & 33.33 & 53.56 & 37.67 & 39.58 & 41.03 \\
\rowcolor{ourrow}
 & C & \textbf{95.41} & \textbf{70.56} & \textbf{48.17} & \textbf{41.67} & \textbf{63.95} \\
\bottomrule
\end{tabularx}
\endgroup
\par\smallskip
\begin{minipage}{\textwidth}\small
Test success (\%) at high reasoning effort. The mean weights all benchmarks equally.
\end{minipage}
\par\normalsize
\caption{Test success by improver and harness edit scope.}
\label{tab:scope-results}
\end{table*}

\subsection*{4.2 RQ2: Improver Performance by Edit Scope}
Which improver benefits most when a proposal can change several components together? Table~\ref{tab:scope-results} crosses the three improvers with all three scopes on each benchmark.

\textbf{Gain by improver.} From A to C, mean success rises by 7.24 points for Luna, 5.75 for Terra, and 25.91 for Sol. Only Sol improves from A to B and again from B to C on all four tasks. From A to C, Luna loses ground on WebShop and Terra on OSBench.

\textbf{Gap between improvers.} Sol leads the next improver by 0.35 points in A, 4.37 in B, and 20.51 in C. The widest scope thus separates the improvers most clearly under the reported training pools, budgets, and selection rules.

\begin{table*}[!t]
\centering
\small
\setlength{\tabcolsep}{2.8pt}
\setlength{\arrayrulewidth}{.35pt}
\arrayrulecolor{black!32}
\renewcommand{\arraystretch}{1.13}
\begin{tabularx}{\textwidth}{@{}l|*{2}{r>{\raggedright\arraybackslash}X|}r>{\raggedright\arraybackslash}X@{}}
\hline
\rowcolor{grouphead}
 & \multicolumn{2}{c|}{\textbf{GPT-6 Luna}} & \multicolumn{2}{c|}{\textbf{GPT-6.1 Terra}} & \multicolumn{2}{c}{\textbf{GPT-6.1 Sol}} \\
\hline
Benchmark & $\Delta$ & Mechanism & $\Delta$ & Mechanism & $\Delta$ & Mechanism \\
\hline
ALFWorld & +32.11 & Training traces show receipt reminders and wrong-object checks. & +36.40 & Training traces show pickup guidance and a target check in two-object tasks. & \textbf{+78.90} & Inventory and search history tell the action check when to require a pickup or stop a repeated search. \\[2pt]
\hline
DBBench & +13.00 & An empty read can hold up submission until the agent checks the result. & +10.00 & A SQL skill tells the agent how to quote identifiers. & \textbf{+26.56} & A log of successful writes guards two decisions: whether to repeat a write and whether to submit. \\[2pt]
\hline
WebShop & +1.17 & A checklist prompts the agent to inspect options and price. & +3.33 & A search skill preserves model numbers and sizes across queries. & \textbf{+18.67} & Search and option plugins share product state. Checkout can be blocked until a required option is selected. \\[2pt]
\hline
OSBench & +14.61 & A counting skill distinguishes files from matching lines. & +12.53 & A command check flags misleading pipelines. Receipt annotations explain the output. & \textbf{+20.17} & A skill specifies which files to inspect and which quantity to count. \\[2pt]
\hline
\end{tabularx}
\par\normalsize
\caption{Mechanisms evolved by each improver under scope C. $\Delta$ is the change in test success (percentage points) relative to bare Qwen3-4B on the same benchmark (Appendix D).}
\label{tab:rq3-improvers}
\end{table*}

\begin{table*}[!t]
\centering
\small
\setlength{\tabcolsep}{2.8pt}
\setlength{\arrayrulewidth}{.35pt}
\arrayrulecolor{black!32}
\renewcommand{\arraystretch}{1.13}
\begin{tabularx}{\textwidth}{@{}l|r>{\raggedright\arraybackslash}X|r>{\raggedright\arraybackslash}X@{}}
\hline
\rowcolor{grouphead}
 & \multicolumn{2}{c|}{\textbf{Scope A $\to$ B}} & \multicolumn{2}{c}{\textbf{Scope B $\to$ C}} \\
\hline
Benchmark & $\Delta$ & Mechanism & $\Delta$ & Mechanism \\
\hline
ALFWorld & +6.12 & Target-tracking instructions gain inventory updates from tool receipts. & \textbf{+62.08} & The action check goes beyond call format: it reads inventory and can block departure before pickup. \\[2pt]
\hline
DBBench & +0.67 & SQL naming guidance is supplemented by repair suggestions after a query fails. & \textbf{+17.00} & A log of successful writes lets the harness block duplicate writes or early submission. \\[2pt]
\hline
WebShop & +3.33 & The option checklist gains page feedback that highlights matching choices. & \textbf{+10.50} & Search and checkout share option state, so a missing selection can block purchase. \\[2pt]
\hline
OSBench & +1.85 & A receipt annotation tells the agent whether its command counted files or lines. & \textbf{+2.08} & A written rule for file selection and aggregation replaces the output annotation. \\[2pt]
\hline
\end{tabularx}
\par\normalsize
\caption{Mechanisms evolved by Sol as edit scope expands. $\Delta$ is the change in test success from the narrower to the wider scope, in percentage points.}
\label{tab:rq3-scopes}
\end{table*}

\subsection*{4.3 RQ3: How Do Improvers and Edit Scopes Shape the Harness?}
We examine the selected harnesses in two comparisons. At fixed scope C, Table~\ref{tab:rq3-improvers} links each improver's gain over the bare executor to the mechanism it produced. With Sol fixed, Table~\ref{tab:rq3-scopes} compares the successive scope gains and traces how the selected mechanisms use tool feedback.

\paragraph{How do gains vary with the mechanism?}
Under scope C, Sol yields the largest gain over the bare executor on every benchmark (Table~\ref{tab:rq3-improvers}). On WebShop, Luna's checklist and Terra's search instructions gain 1.17 and 3.33 points, respectively. Sol gains 18.67 points with a harness that records selected options and consults that record before checkout.

\paragraph{Which dependencies matter?}
On DBBench, both Luna and Sol act on query history. Luna asks for a second look after an empty result. Sol uses the outcome of a write to govern both a repeat attempt and submission. OSBench shows a different ordering: Terra uses both a command check and receipt annotations, while Sol obtains the highest score with a written counting rule.

\noindent\textbf{How much does joint editing add beyond independent component edits?} With Sol fixed, we compare two differences in test success on each benchmark. A edits skill text. B can revise one component per proposal. C can revise how one handler records evidence and how another uses it in the same proposal.

\paragraph{Where do independent edits fall short?}
Across benchmarks, mean success gains are 2.99 points from A to B and 22.92 from B to C. A-to-B gains range from 0.67 to 6.12 points (Table~\ref{tab:rq3-scopes}). B can already improve the information supplied to the agent. On ALFWorld it even has both inventory memory and an action check, but the check validates only call format. C connects those components: inventory now determines whether the harness permits the next move. The gain from B to C is 62.08 points, compared with 6.12 from A to B.

\paragraph{When is the second gain larger?}
DBBench and WebShop show the same pattern. Once a write receipt or an option selection becomes a condition on a later action, the B-to-C gain exceeds the A-to-B gain by 16.33 and 7.17 points, respectively. OSBench provides a contrasting case: C retains a counting instruction, and the two gains differ by only 0.23 points. Thus the largest increases coincide with harnesses that use earlier observations to control later actions. The gains compare selected harnesses from separate searches, including a continued search for WebShop C.

\section*{5 Conclusion}
HookEvo lets an improver revise the handlers that record tool feedback and those that use it in a single candidate. With a frozen Qwen3-4B executor, the selected Sol-C harnesses achieved the highest measured test means among the reported Qwen3-4B comparisons on all four benchmarks. Moving from skill-text edits to coordinated edits yielded different gains across improvers: 25.91 percentage points for Sol, 7.24 for Luna, and 5.75 for Terra on average. The retained harnesses also differed in how they used that scope. On WebShop, Sol linked option selection to a checkout check, while on OSBench its retained harness supplied a written counting rule.

\paragraph{Limitations.}
Training data allocation varied across some runs: some reused a fixed task pool, while others rotated through batches. These schedules expose the improver to different trajectories, so the contribution of data allocation to the observed performance differences remains unclear.

\paragraph{Next steps.}
First, we will test whether an improver can use execution traces to choose the edit scope for each proposal under a fixed search budget. Second, we will disable the handlers that write and read episode state, separately and together, to measure how their interaction affects task success.

\bibliography{references}
\section*{A. Harness Evolution Algorithm}
\label{app:algorithm}
The A/B/C controllers implement the architecture in Figure~\ref{fig:method-overview}: observe the current harness, propose a scoped edit, compare the candidate with its parent, and feed the measured decision into the next update. The editable scope determines the candidate representation. The executor, runtime hooks, and evaluation controller provide the surrounding execution structure.

\subsection*{A.1 Harness State and Runtime Interfaces}
A harness $H=(\lambda,G)$ consists of skill instructions $\lambda$ and an executable composition $G$. The runtime creates fresh state for each episode and invokes handlers at the corresponding execution events. An observation handler reads a tool call and its receipt; a guidance handler exposes relevant state in the next model context; an action handler checks a proposed call; and an explanation handler annotates tool feedback. These interfaces let an edit change how evidence is acquired, retained, and used in subsequent decisions.

Scope A replaces skill text. Scope B replaces one independent component: skill text, memory, an action check, or receipt interpretation. Unedited components are inherited, and executable components maintain separate state. Scope C replaces skill text or a composition of named mechanisms whose handlers can share episode state. A coordinated edit can therefore introduce a fact, its producer, and its downstream consumers together.

\subsection*{A.2 Diagnosis and Candidate Construction}
The improver receives the current instructions and executable source, the available runtime interfaces, sampled successful and failed trajectories, and the preceding evaluation decision. The source and interfaces describe the parent's existing capabilities. The trajectories reveal where those capabilities fail or succeed; the preceding decision supplies the measured effects of the last attempted repair.

A proposal identifies the editable aspect, supplies its replacement, and explains the evidence and intended behavioral change. Diagnosis and implementation are returned together. For a composed mechanism, the proposal specifies which handler records each fact and which handlers consume it. The host applies the replacement to a copy of the parent and checks the replacement format and executable syntax before evaluation.

\subsection*{A.3 Evolution Procedure}
Algorithm~\ref{alg:harness} summarizes one search from initialization through held-out evaluation.

\begin{algorithm}[tb]
\caption{Feedback-Guided Harness Evolution}
\label{alg:harness}
\textbf{Input:} initial harness $H_0$, frozen executor, improver $M$, edit scope $k$, training schedule, optional validation set, and search budget $B$.\par
\textbf{Initialize:} retained harness $H\leftarrow H_0$; previous feedback $\mathcal R\leftarrow\varnothing$.\par
\begin{enumerate}
\item \textbf{Observe.} While budget remains, obtain the next training batch and the parent outcomes, trajectories, and hook outputs. Select successful and failed traces as proposal evidence.
\item \textbf{Attribute and propose.} Give $M$ the parent source, runtime interfaces, sampled trajectories, previous feedback $\mathcal R$, and allowed edits $\mathcal E_k(H)$. Obtain a diagnosis and one candidate edit.
\item \textbf{Construct the candidate.} Copy the parent and apply the scoped replacement to form $H'$. If it cannot be assembled, retain $H$ and pass the assembly result to the next round.
\item \textbf{Execute and compare.} Run $H'$ through the runtime hooks with fresh episode state. Pair its results with the parent outcomes on matching tasks, trials, and seeds. Compare the training objective, emitted hook outputs, repaired failures, and regressions. Compare validation scores when a validation set is present.
\item \textbf{Select.} Retain $H'$ if its training objective improves, its emitted hook outputs change, and its validation objective does not decrease when used. Otherwise retain $H$.
\item \textbf{Feed back.} Set $\mathcal R$ to this round's proposal, measured outcomes, and retention decision. Continue from the retained harness.
\end{enumerate}
\textbf{Output:} the retained harness and the sequence of evaluated edits and decisions. Evaluate retained checkpoints on held-out tasks after search.\par
\end{algorithm}

\subsection*{A.4 Selection and Feedback}
Let $J_{\mathrm{tr}}$ denote the configured training objective, and let $a(H',H)$ indicate a difference in the emitted hook outputs on paired tasks. The retention rule is
\[
\begin{aligned}
\operatorname{Accept}(H',H)&=
\big[J_{\mathrm{tr}}(H')>J_{\mathrm{tr}}(H)\big]
\\
&\quad\land a(H',H)\land v(H',H),
\end{aligned}
\]
where $v$ is true if no validation set is used, and otherwise requires $J_{\mathrm{val}}(H')\geq J_{\mathrm{val}}(H)$. Hook outputs comprise context guidance, action-check reasons, and receipt annotations; plugin-load events are excluded. The next proposal receives the latest decision, including both repairs and regressions. Accepted code persists across rounds, while runtime state is reset between episodes.

\subsection*{A.5 Correspondence to the A/B/C Implementation}
The bundled controllers implement observation with \texttt{score} and \texttt{trace\_sample}; \texttt{model} assembles the source, interface, sampled evidence, and previous feedback into the improver request. Its response binds the rationale to the replacement through the fields \texttt{reason}, \texttt{aspect}, and \texttt{content}. The \texttt{install} function realizes the scope-specific replacement. These operations correspond to the evidence inputs, improver workspace, and candidate block in Figure~\ref{fig:method-overview}.

The runtime wrappers implement the diagram's execution block: \texttt{guide} contributes to context assembly, \texttt{check} runs before tool execution, and \texttt{observe} followed by \texttt{explain} processes the receipt. B constructs independent memory, action, and receipt components; C calls \texttt{create(state)} to construct mechanisms sharing fresh episode state. The controllers then use \texttt{measure} and \texttt{signature} for paired scores and observable hook effects, record the retention decision, and provide it to the next \texttt{model} call. This closes the diagram's retained-code and measured-feedback loop.

\subsection*{A.6 Improver Prompt and Scope-Specific Instructions}
The following template reproduces the instruction text in the controllers' \texttt{model} function. Angle-bracket placeholders replace the values inserted at each round. The same template is used for A, B, and C; the editable-aspect descriptions and supplied executable sources depend on scope.

\begin{quote}\raggedright
Role: improve a frozen Qwen agent in \textless{}benchmark\textgreater{} by turning training observations into one reusable candidate edit.
\par
Inputs: the current parent, visible training traces, previous measured feedback, and the runtime interfaces.
\par
Editable aspects: \textless{}editable-aspect descriptions\textgreater{}.
\par
Choose exactly one aspect. Return its complete replacement in content and describe the evidence, expected behavior, and potential regression in reason.
\par
All other aspects are inherited. Each executable aspect is a self-contained JavaScript module, uses only its function arguments and its own local state, and is wired to its designated interface by the host.
\par
The host runs paired evaluation, records observed effects and decides promotion. The traces below are observations.
\par
Parent: \textless{}parent identifier\textgreater{}
\par
Interface reference: \textless{}runtime interface reference\textgreater{}
\par
Current skill: \textless{}parent skill text\textgreater{}
\par
Current executable aspects: \textless{}parent executable sources\textgreater{}
\par
Previous measured feedback: \textless{}previous decision\textgreater{}
\par
Training observations: \textless{}sampled training traces\textgreater{}
\end{quote}

\paragraph{Editable-aspect descriptions.}
The controller inserts the following descriptions as a JSON object. A includes only \texttt{skill}; B includes \texttt{skill}, \texttt{memory}, \texttt{action}, and \texttt{receipt}; C includes \texttt{skill} and \texttt{composition}.

\noindent\textbf{skill.} Reusable strategy and task guidance as plain text.\par

\noindent\textbf{memory.} Episode-local state. JS export function create() returns \{observe(call, receipt), guide()\}. observe updates local facts; guide returns a short factual state summary.\par

\noindent\textbf{action.} One action precondition. JS export function create() returns \{check(call)\}. Return a reason string to ask the actor to revise, or undefined to continue.\par

\noindent\textbf{receipt.} One receipt interpretation or recovery rule. JS export function create() returns \{explain(call, receipt)\}. Return a short annotation or undefined.\par

\noindent\textbf{composition.} One coherent executable mechanism edit. Return JSON text in content with keys skill and code. code is a JavaScript module exporting create(state), returning an ordered array of named mechanisms \{id, observe(call, receipt), guide(), check(call), explain(call, receipt)\}. Handlers are optional. The host supplies episode-local shared state. Name which mechanism writes each fact and which other mechanism uses it. All mechanisms may be edited together when the proposed repair requires it.\par

For C, ``own local state'' in the common prompt is the episode-local state supplied to the composition; its mechanisms share that object as specified in the composition instruction. The controller supplies the parent source and runtime interfaces directly to the improver in Figure~\ref{fig:method-overview}.

\paragraph{Evidence inserted into the prompt.}
Each sampled trajectory supplies its task identifier, native score, and opening and closing messages. Sampling draws from both successful and failed parent episodes, using the round identifier to seed the sampler. The previous-decision field is empty initially and thereafter contains the latest measured decision. The exact rendered prompt is saved before the improver call, so a proposal can be replayed with its original evidence. The bundled sampler specifies serialization and truncation details.

\subsection*{A.7 Candidate and Feedback Formats}
The improver returns one JSON object. All three fields below are required strings, and the response schema permits no additional fields:
{\small
\begin{verbatim}
{
  "aspect": "<permitted aspect>",
  "content": "<complete replacement>",
  "reason": "<rationale>"
}
\end{verbatim}
}
The permitted aspects are those listed in A.6. For a skill edit, \texttt{content} is the replacement text. For B's executable aspects, it is module source exporting \texttt{create()}. For C's composition edit, it is a serialized JSON object with two fields, \texttt{skill} and \texttt{code}. The latter exports \texttt{create(state)} and returns an ordered array of named mechanisms. The following is an interface skeleton, with handler bodies supplied by the improver:
{\small
\begin{verbatim}
export function create(state) {
  return [{
    id: "mechanism-name",
    observe(call, receipt) {
      /* update state */ },
    guide() { /* context guidance */ },
    check(call) {
      /* denial or undefined */ },
    explain(call, receipt) {
      /* annotate receipt */ }
  }];
}
\end{verbatim}
}
Handlers are optional. The host calls observation handlers before receipt-explanation handlers, adds guidance during context assembly, and applies action checks before native tool execution. This ordering implements the information path shown in Figure~\ref{fig:method-overview}.

The feedback record contains candidate and parent identifiers, the batch, the edited aspect and proposal rationale, before/after measurements, optional validation measurements, the observed-effect flag, repaired and regressed task counts, and the adoption decision. Measurements include strict success and mean native reward; the configured objective determines promotion. Invalid assembly instead records the error and retains the parent. Only the latest decision is inserted into the next prompt; the sequence of decisions is saved separately.

\section*{B. Ablation Setup for Candidate Edit Scope}
\label{app:configurations}
\paragraph{Executor and environments.}
We use a frozen Qwen3-4B-Instruct-2507 executor on ALFWorld, DBBench, WebShop, and OSBench. Each benchmark adapter supplies task initialization, native tools, environment feedback, and scoring. Training trajectories drive candidate generation and selection. Validation, where present, supplies the non-degradation check in Appendix~A. Held-out evaluation follows the completed search.

\paragraph{Improvers and editable scopes.}
The study crosses GPT-6 Luna, GPT-6.1 Terra, and GPT-6.1 Sol, each at high reasoning effort, with scopes A, B, and C. The scope definitions below govern candidate construction within the common loop in Appendix~A.

\noindent\begin{minipage}{\columnwidth}
\centering\small
\begin{tabularx}{\linewidth}{@{}lX@{}}
\toprule
Scope & Permitted update\\
\midrule
A & Replace skill instructions.\\
B & Replace one independent skill, memory, action-check, or receipt component; inherit the others.\\
C & Replace skill instructions or jointly revise executable mechanisms and their shared episode state.\\
\bottomrule
\end{tabularx}
\par\normalsize
\captionof{table}{Editable representations in the A/B/C study.}
\label{tab:scope-settings}
\end{minipage}

\paragraph{Training batches and search budgets.}
In rotating-pool runs, we partition 300 distinct training tasks into six batches of 50 and cycle through the batches. Each round uses one batch to generate evidence for one proposed harness update and to compare the candidate with its parent. We limit the evidence considered at each update because long contexts can make it harder for the improver to focus on the steps relevant to a failure. Fixed-pool runs reuse their training pool each round. Most runs in the scope comparison have a 40-round budget; the ALFWorld Sol-C result in Table~\ref{tab:scope-results} comes from a 20-round run. Figure~\ref{fig:evolution} shows the histories associated with the leaderboard checkpoints: 40 rounds each for ALFWorld, DBBench, and OSBench, and two successive 40-round stages for WebShop.

\paragraph{Evaluation procedure.}
Within each update, parent and candidate are compared on the same task, trial, and seed schedule under the benchmark's training objective. The controller records score changes and observed execution effects before making the retention decision. The task manifests and objective are inputs to this procedure; the edit scope determines which replacements the improver may propose.

\paragraph{Reported measurements.}
The held-out sets contain 109 ALFWorld, 300 DBBench, 200 WebShop, and 144 OSBench tasks. Results report native task success. WebShop counts fully successful purchases. Headline evaluations use three executions. Table~\ref{tab:scope-results} reports the mean task success for each improver and edit scope. Appendix~D defines checkpoint reporting and the evolution curves. Appendix~E describes the resulting harness mechanisms.
\begin{table*}[!tp]
\section*{C. Additional Method and Model Comparisons}
\centering\small
\setlength{\tabcolsep}{2pt}
\begin{tabularx}{\textwidth}{@{}>{\raggedright\arraybackslash}p{.148\textwidth}>{\raggedright\arraybackslash}p{.212\textwidth}*{4}{>{\centering\arraybackslash}X}>{\raggedright\arraybackslash}p{.25\textwidth}@{}}\toprule
Executor & Method / harness & ALFWorld & DBBench & WebShop & OSBench & Source\\\midrule
\rowcolor{grouphead}\multicolumn{7}{l}{\textbf{Qwen3-4B: evolved artifacts and additional baselines}}\\
\rowcolor{ourrow}Qwen3-4B & Ours: Sol high & 95.41$\pm$0.00 & 70.56$\pm$0.38 & 48.17$\pm$0.29 & 41.67$\pm$1.84 & Local\\
Qwen3-4B & ALIGN (released)\citep{align} & 3.36$\pm$1.06 & -- & 31.00$\pm$0.50 & -- & Local\\
Qwen3-4B & ReAct\citep{react} & 10.70$\pm$1.06 & -- & -- & -- & Local\\
Qwen3-4B & ReflAct\citep{reflact} & 18.35$\pm$1.83 & -- & -- & -- & Local\\
Qwen3-4B & AutoManual\citep{automanual} & 76.15$\pm$1.59 & -- & -- & -- & Local\\
Qwen3-4B & Harness-R1 patch\citep{harness-r1} & -- & 52.22$\pm$0.69 & -- & -- & Local\\
Qwen3-4B & Prompt-evolving$^{\dagger}$ & 13.8 & 60.7 & 30.0 & 18.8 & \citep{life}\\
Qwen3-4B & SkillFM & 77.01 & -- & 14.60 & -- & \citep{skillfm}\\
\addlinespace[3pt]
\rowcolor{grouphead}\multicolumn{7}{l}{\textbf{Larger open-model executors}}\\
Qwen2.5-14B & Life-Harness & 83.49 & 66.00 & 45.00 & 40.97 & \citep{life}\\
Qwen2.5-32B & Life-Harness & 92.66 & 69.33 & 46.50 & 45.83 & \citep{life}\\
Qwen2.5-72B & Life-Harness & 85.32 & 71.33 & 48.00 & 47.22 & \citep{life}\\
Llama-3.3-70B & Life-Harness & 92.66 & 68.67 & 44.00 & 39.58 & \citep{life}\\
Qwen3.6-27B & Life-Harness & 89.91 & 74.33 & 43.00 & 56.94 & \citep{life}\\
Qwen3.6-35B-A3B & Life-Harness & 89.91 & 71.33 & 45.00 & 54.17 & \citep{life}\\
\addlinespace[3pt]
\rowcolor{grouphead}\multicolumn{7}{l}{\textbf{Frontier-model executors}}\\
GPT-5.5 high & Bare & 93.6 & 67.7 & 75.1 & 47.4 & \citep{life}\\
GPT-5.5 high & Life-Harness & 95.4 & 75.0 & 75.8 & 53.1 & \citep{life}\\
Opus 4.8 high & Bare & 97.2 & 71.7 & 73.8 & 50.4 & \citep{life}\\
Opus 4.8 high & Life-Harness & 98.1 & 74.3 & 82.7 & 50.8 & \citep{life}\\
Gemini 3.1 Pro & Bare & 96.3 & 76.5 & 67.3 & 50.0 & \citep{life}\\
Gemini 3.1 Pro & Life-Harness & 96.3 & 78.4 & 68.1 & 48.6 & \citep{life}\\
\addlinespace[3pt]
\rowcolor{grouphead}\multicolumn{7}{l}{\textbf{Additional optimization methods}}\\
GPT-5.5 & TextGrad & 82.80 & -- & -- & -- & \citep{textgrad,skillopt}\\
GPT-5.5 & GEPA & 85.80 & -- & -- & -- & \citep{gepa,skillopt}\\
GPT-5.5 & SkillOpt & 95.50 & -- & -- & -- & \citep{skillopt}\\
GPT-4o & AgentSquare & 69.50 & -- & -- & -- & \citep{agentsquare}\\
GPT-4o & ADAS & 54.30 & -- & -- & -- & \citep{adas,agentsquare}\\
GPT-4o & ReflAct & 93.30 & -- & -- & -- & \citep{reflact}\\
GPT-4-turbo & AutoManual & 97.40 & -- & -- & -- & \citep{automanual}\\
GPT-3.5 series & AutoGuide & 79.10 & -- & 46.00 & -- & \citep{autoguide}\\
PaLM-540B & ReAct & 57.00 & -- & 40.00 & -- & \citep{react}\\
\bottomrule\end{tabularx}
\smallskip\begin{minipage}{\textwidth}\small
The evolved Qwen3-4B harnesses exceed the displayed Life-Harness Qwen2.5-14B scores on all four benchmarks and the Qwen2.5-32B scores on ALFWorld, DBBench, and WebShop.

\smallskip
$^{\dagger}$Prompt-evolving values are approximate readings from Life-Harness Figure 6. SkillFM retains its published task counts and protocol. Life-Harness Tables 10 and 2 provide the larger open-model and frontier-model rows.
\end{minipage}
\par\normalsize
\caption{Comparison of evolved Qwen3-4B harnesses with additional methods and larger executors. Scores are test success (\%); local measurements report mean and sample SD. Sources identify the numerical evaluations.}
\label{tab:cross-model-context}
\end{table*}

\FloatBarrier

\section*{D. Metric Definitions and Reporting}
\label{app:metrics}
\paragraph{Test success and repeated execution.}
For $N$ held-out tasks and repeat $r$, let $s_{ir}\in\{0,1\}$ denote native task success. The repeat-level percentage is $p_r=100N^{-1}\sum_{i=1}^{N}s_{ir}$. We report the mean over the available complete repeats and their sample standard deviation:
\[
\bar p=\frac{1}{R}\sum_{r=1}^{R}p_r,\qquad
\mathrm{SD}=\sqrt{\frac{\sum_{r=1}^{R}(p_r-\bar p)^2}{R-1}}.
\]
Local headline evaluations use three repeats. Table~\ref{tab:scope-results} reports success rates only. Luna and Terra under C on DBBench and OSBench each use one execution. WebShop success counts fully successful purchases.

\paragraph{Baseline-adjusted gains and error bars.}
Gain is measured test success minus the Life-Harness bare-Qwen3-4B reference~\citep{life}: 16.51\%, 44.00\%, 29.50\%, and 21.50\% for ALFWorld, DBBench, WebShop, and OSBench. Error bars show the sample SD of repeated checkpoint evaluations.

\paragraph{Evolution traces.}
Candidate training includes each recorded proposal. Retained training uses the accepted candidate's score, or its evaluated parent's score if the proposal is rejected, on that round's task batch. Test markers identify measured checkpoints. The horizontal axis spans the full search budget. Figure~\ref{fig:evolution} averages available training scores over the latest six search rounds for the rotating-batch panels; it preserves the initial measured point and leaves the fixed-pool ALFWorld and OSBench scores unaveraged. The window resets at the WebShop stage boundary. Early rounds average the available recorded scores.

\begin{figure*}[!tp]
\section*{E. Harness Architectures and Implementation Details}
\label{app:artifact-details}
\subsection*{E.1 ALFWorld}
\centering
\includegraphics[width=\textwidth]{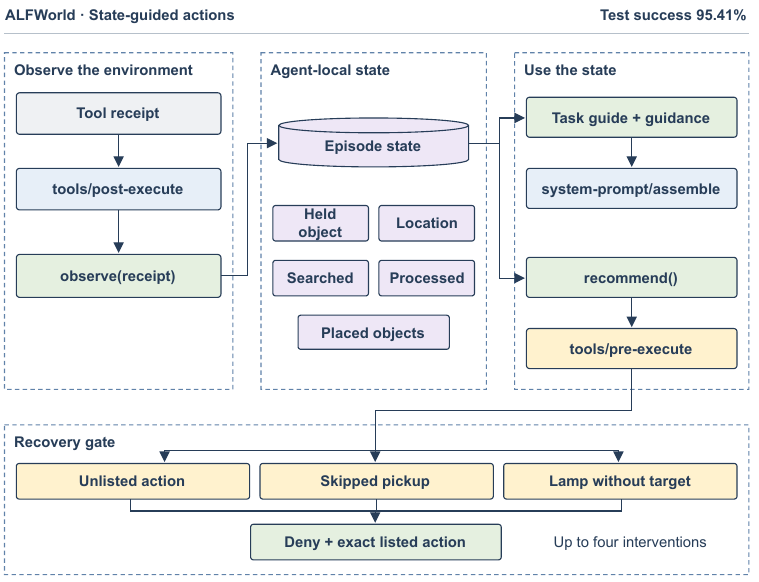}
\caption{ALFWorld harness. The action-hint plugin parses accepted tool receipts and records episode state. Context assembly and the pre-execution guard use that state. The task-guide plugin supplies the household procedure.}
\label{fig:artifact-alfworld-detail}
\end{figure*}
\begin{figure*}[!tp]
\subsection*{E.2 DBBench}
\centering
\includegraphics[width=\textwidth]{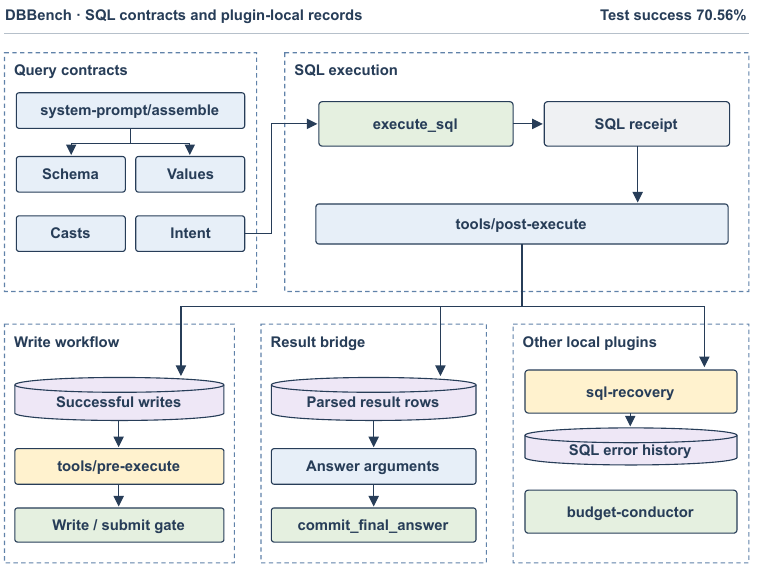}
\caption{DBBench harness. Query contracts constrain SQL construction. The write-workflow, result-bridge, sql-recovery, and budget-conductor plugins maintain local records. The result bridge parses SQL receipts into final-answer arguments.}
\label{fig:artifact-dbbench-detail}
\end{figure*}
\begin{figure*}[!tp]
\subsection*{E.3 WebShop}
\centering
\includegraphics[width=\textwidth]{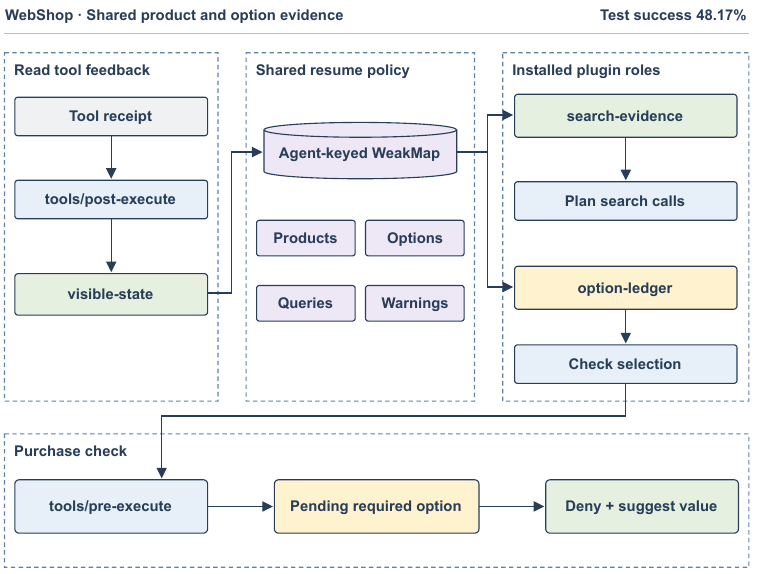}
\caption{WebShop harness. The search-evidence and option-ledger plugins use the same resume policy and share an agent-keyed WeakMap. Receipt observations update product and option evidence used in later search and purchase checks.}
\label{fig:artifact-webshop-detail}
\end{figure*}
\begin{figure*}[!tp]
\subsection*{E.4 OSBench}
\centering
\includegraphics[width=\textwidth]{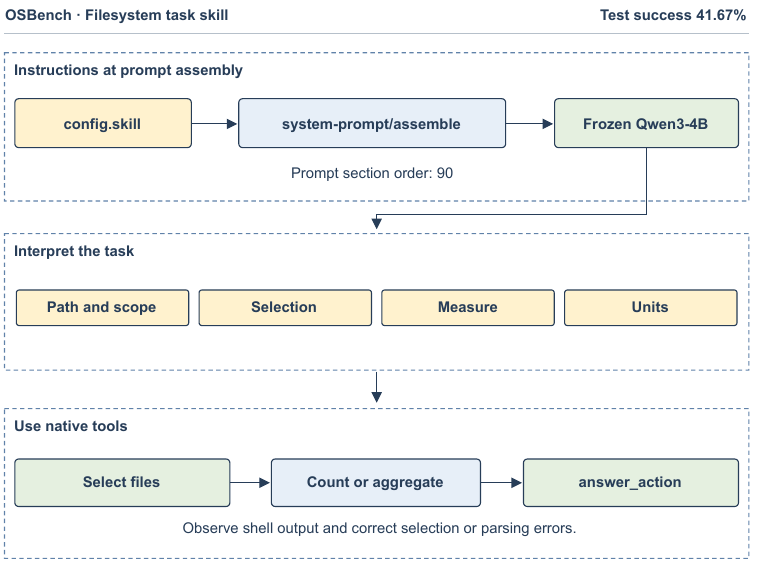}
\caption{OSBench harness. The retained skill in config.skill is added at prompt assembly. The executable composition is empty. The frozen executor uses the native shell tools and submits through the native answer action.}
\label{fig:artifact-osbench-detail}
\end{figure*}

\FloatBarrier
\subsection*{E.5 Core Implementation Excerpts}
These excerpts show the main state updates, decisions, or hook bindings of each installed plugin in Figures~\ref{fig:artifact-alfworld-detail}--\ref{fig:artifact-osbench-detail}. The WebShop helper modules and the OSBench composition are included because they explain how state and skills are used. Comments mark omissions, and long lines wrap for display.

\lstset{
  basicstyle=\footnotesize\ttfamily,
  showstringspaces=false,
  columns=fullflexible,
  keepspaces=true,
  breaklines=true,
  breakatwhitespace=false,
  tabsize=2,
  literate={×}{{\texttimes}}1 {–}{{\textendash}}1 {—}{{\textemdash}}1
           {“}{{\textquotedblleft}}1 {”}{{\textquotedblright}}1
           {→}{{\ensuremath{\rightarrow}}}1 {−}{{\ensuremath{-}}}1
}

\subsubsection*{ALFWorld}\mbox{}\par

\paragraph{\texttt{task-guide.mjs}.} Adds the household procedure to the assembled system prompt.
\begin{lstlisting}
    ac.on('system-prompt/assemble', async (assembly, context, next) => {
// ...
      return {...result,sections:[...result.sections,{name,order:25,text,interpolate:false}]};
\end{lstlisting}

\paragraph{\texttt{action-hint.mjs}.} Reads accepted receipts into episode state and checks the next action against the current task facts. The guard uses the state to recommend a recovery action.
\begin{lstlisting}
    ac.on('tools/post-execute', async (call, result, next) => {
      const decision = await next();
      if (decision.kind !== 'accept') return decision;
      const raw = getRawBusinessResult(call.callId);
      observe(typeof raw?.content === 'string' ? raw.content : textOf(result.content));
// ...
      const recovery = recommend();
      const skippedPickup = !held && recovery?.startsWith('take ' + target + ' ') && action !== recovery;
      const emptyLamp = lamp && String(action).startsWith('use ' + lamp + ' ') && held.split(' ')[0] !== target;
      if (recovery && (!actions.includes(action) || skippedPickup || emptyLamp)) {
        recoveries++;
\end{lstlisting}

\subsubsection*{DBBench}\mbox{}\par

\paragraph{\texttt{schema-contract.mjs}.} Uses the exact schema identifiers and checks reserved column names before SQL execution.
\begin{lstlisting}
  const t=p.taskContext?.table||{}, columns=(t.table_info?.columns||[]).map(x=>x.name);
  const quote=s=>'`'+String(s).replaceAll('`','``')+'`';
// ...
  if(config.guard)p.ac.on('tools/pre-execute',async(call,next)=>{
// ...
   const bad=columns.filter(x=>/^rank$|^group$|^order$/i.test(x)).find(x=>new RegExp('\\b'+x+'\\b','i').test(masked));
\end{lstlisting}

\paragraph{\texttt{literal-contract.mjs}.} Collects example cell values and exposes guidance on exact literal matching at prompt assembly.
\begin{lstlisting}
  const samples=cols.map((col,i)=>({column:col.name,examples:[...new Set(rows.map(r=>String(r[i]??'')))].slice(0,config.samples||3)}));
// ...
  p.ac.on('system-prompt/assemble',async(a,c,next)=>{
   const r=await next();p.audit({method_id:name,hook:'system-prompt/assemble',text});return {...r,sections:[...r.sections,{name,order:30,text,interpolate:false}]};
\end{lstlisting}

\paragraph{\texttt{write-workflow.mjs}.} Records successful writes and attaches checks for duplicate writes and premature submission.
\begin{lstlisting}
  p.ac.on('tools/post-execute',async(call,result,next)=>{
// ...
   if(/^(INSERT|UPDATE|DELETE)\b/i.test(q.trim())&&!/^\d{4} \(/.test(raw)&&!p.getRawBusinessResult(call.callId)?.error){writes.push(q);checked=false;}
// ...
  if(config.dedupe)p.ac.on('tools/pre-execute',async(call,next)=>{
// ...
  if(config.proof)p.ac.on('tools/pre-execute',async(call,next)=>{
\end{lstlisting}

\paragraph{\texttt{result-bridge.mjs}.} Parses SELECT receipts into values or tuples for final-answer arguments.
\begin{lstlisting}
   if(!/^SELECT\b/i.test(args?.query?.trim()||''))return d;
   latest=raw;const rows=rowsOf(raw);answers=rows.map(row=>{const cells=cellsOf(row);return cells.length===1?decode(cells[0]):row;});
   if(!config.annotate||!rows.length)return d;
\end{lstlisting}

\paragraph{\texttt{sql-recovery.mjs}.} Keeps a local record of SQL errors and counts repeated failures.
\begin{lstlisting}
   const q=args?.query||'',raw=p.getRawBusinessResult(call.callId)?.content||'';let annotation='';
   if(/^\d{4} \(/.test(raw)){
    last=q;errors.push(raw.slice(0,300));repeats.set(q,(repeats.get(q)||0)+1);
\end{lstlisting}

\paragraph{\texttt{budget-conductor.mjs}.} Tracks SQL calls, writes and errors, and obtains the current step for budget guidance.
\begin{lstlisting}
   const q=args?.query||'',raw=p.getRawBusinessResult(call.callId)?.content||'';last=q;
   if(/^\d{4} \(/.test(raw))errors++;else if(/^(INSERT|UPDATE|DELETE)\b/i.test(q.trim()))write++;
// ...
  p.ac.on('system-prompt/assemble',async(a,c,next)=>{
   const r=await next();const step=p.getVisibleInput()?.step||1;
\end{lstlisting}

\paragraph{\texttt{numeric-calendar.mjs}.} Uses a pre-execution check to catch numeric ordering and aggregation over text columns.
\begin{lstlisting}
   p.ac.on('tools/pre-execute',async(call,next)=>{
// ...
     if(new RegExp('\\b(?:MIN|MAX)\\(\\s*'+token+'\\s*\\)','i').test(q)||new RegExp('\\bORDER\\s+BY\\s+'+token+'(?:\\s+(?:ASC|DESC))?(?:\\s+LIMIT|\\s*;?\\s*$|\\s*,)','i').test(q)){
\end{lstlisting}

\paragraph{\texttt{intent-contract.mjs}.} Separates fields used to select a row from fields used to assign new values.
\begin{lstlisting}
  const selector=m&&field(m[1]),target=m&&field(m[3]);
  const plan=selector&&target?{selector:{column:selector,value:m[2]},assignment:{column:target,value:m[4]}}:null;
\end{lstlisting}

\subsubsection*{WebShop}\mbox{}\par

\paragraph{\texttt{search-evidence.mjs}.} Installs the shared policy and refreshes the visible page and price bound from tool receipts.
\begin{lstlisting}
 installResumeRole(ctx,name,resumeFeatures);
// ...
   const previous=latest;const before=actions(previous);latest=raw;
   const obs=raw.split('Available Actions:')[0];
   const instruction=obs.match(/Instruction:\s*\[SEP\]\s*(.*?)\[SEP\]/s)?.[1]||'';
   const bound=instruction.match(/price lower than\s*([\d.]+)/i);if(bound)upper=Number(bound[1]);
\end{lstlisting}

\paragraph{\texttt{option-ledger.mjs}.} Installs the same policy and checks for blocking option issues at checkout.
\begin{lstlisting}
 installResumeRole(ctx,name,resumeFeatures);
// ...
  ac.on('tools/pre-execute',async(call,next)=>{
// ...
   if(call.name==='click_action'&&norm(args.value)==='buy now'&&(blockingIssues.length||fortuneTypeIssue||flashCapacityIssue||truffleTypeIssue)&&!warned.has(currentId)&&warningCount<2){
\end{lstlisting}

\paragraph{\texttt{resume-policy.mjs}.} Reuses task-local state keyed by agent context and updates it from accepted tool receipts.
\begin{lstlisting}
const states=new WeakMap();
// ...
let st=states.get(ac);
// ...
  ac.on('tools/post-execute',async(call,result,next)=>{
   const d=await next();if(d.kind!=='accept')return d;const receipt=getRawBusinessResult(call.callId);if(!receipt)return d;
   const f=observe(st,call,result,receipt,Number(getVisibleInput().step)||st.step+1);const notes=[];
\end{lstlisting}

\paragraph{\texttt{visible-state.mjs}.} Extracts option groups by matching page values against the visible clickable actions.
\begin{lstlisting}
 const clickable=new Set(actions(raw));const groups=[];let group;
 for(let i=start;i<end-1;i++){
  if(!clickable.has(parts[i])){group={label:parts[i],values:[]};groups.push(group);}
  else if(group)group.values.push(parts[i]);
 }
 return groups.filter(g=>g.values.length);
\end{lstlisting}

\subsubsection*{OSBench}\mbox{}\par

\paragraph{\texttt{scale-slot.mjs}.} Adds the retained filesystem instructions at prompt assembly. The wrapper also accepts guidance from executable mechanisms when any are installed.
\begin{lstlisting}
    ac.on('system-prompt/assemble', async (assembly, context, onward) => {
      const result = await onward();
      const guidance = mechanisms.map(m => m.guide?.()).filter(Boolean);
      const text = [config.skill, ...guidance].filter(Boolean).join('\n');
      if (!text) return result;
\end{lstlisting}

\paragraph{\texttt{scale-composition.mjs}.} Returns an empty mechanism list for the retained harness.
\begin{lstlisting}
export function create() { return []; }
\end{lstlisting}

\paragraph{Retained skill.}
The instructions are stored in \texttt{config.skill} in \texttt{composition.json}. They specify the path and scope, selection rules, target quantity and units. The unselected memory, action and receipt modules are empty stubs.

\end{document}